\pdfoutput=1

\documentclass[11pt]{article}

\usepackage[]{acl} %

\usepackage{times}
\usepackage{latexsym}

\usepackage[T1]{fontenc}

\usepackage[utf8]{inputenc}

\usepackage{microtype}

\usepackage{inconsolata}
\usepackage{enumitem}
\usepackage{amsmath}
\usepackage{amsfonts}
\usepackage{subcaption}
\usepackage{multirow}
\usepackage{graphicx,tabularx}
\usepackage{caption}
\usepackage{booktabs}
\usepackage{algorithm}
\usepackage[noend]{algpseudocode}
\usepackage{algorithmicx}
\usepackage{xcolor}
\usepackage{newfloat}
\usepackage{listings}
\usepackage{url}
\usepackage{makecell}
\usepackage{pifont}
\usepackage{tcolorbox}

\usepackage{xcolor,colortbl}
\usepackage{tabularray}

\title{Persona is a Double-edged Sword: Mitigating the Negative Impact of Role-playing Prompts in Zero-shot Reasoning Tasks}

\author{Junseok Kim$^{1}$, Nakyeong Yang$^{2}$ and Kyomin Jung$^{2}$ \\
  $^{1}$Pohang University of Science and Technology,
  $^{2}$Seoul National University
  \\
  \texttt{junseokkim00@postech.ac.kr} \\
  \texttt{\{yny0506, kjung\}@snu.ac.kr}
  }

\begin{document}
\maketitle


\begin{abstract}
Recent studies demonstrate that prompting a role-playing persona to an LLM improves reasoning capability.
However, assigning an adequate persona is difficult since LLMs are extremely sensitive to assigned prompts; thus, inaccurately defined personas sometimes hinder LLMs and degrade their reasoning capabilities.
In this paper, we first investigate the potential negative impact of injecting persona into language models.
Furthermore, we propose a novel framework, Jekyll \& Hyde, which ensembles the outcomes of both role-playing and neutral prompts to enhance the robustness of reasoning ability.
Specifically, Jekyll \& Hyde predicts an appropriate persona using an LLM when defining the role-playing prompt.
Then, Jekyll \& Hyde collects two potential solutions from role-playing and neutral prompts and selects a better solution using the LLM evaluator.
The experimental analysis demonstrates that role-playing prompts sometimes distract LLMs, degrading their reasoning abilities in 7 out of 12 datasets in llama3. 
Meanwhile, Jekyll \& Hyde improve reasoning capabilities by selecting better choices among the potential solutions on twelve widely-used natural language reasoning datasets. 
In addition, we reveal that assigning LLM-generated personas obtains more stable results than handcrafted personas.
\end{abstract}

\section{Introduction}
Recent studies have exhibited that assigning specific roles to prompts can activate the role-playing ability of Large Language Models (LLMs), improving their reasoning capabilities \cite{shanahan2023role}.
Specifically, some studies have proposed utilizing a hand-crafted persona or analyzing various jobs and relationships to find the most optimal persona that enhances the model's reasoning ability \citep{kong2024better, zheng2023helpful}.

\begin{figure}[t]
\centering
\includegraphics[width=1.0\linewidth]{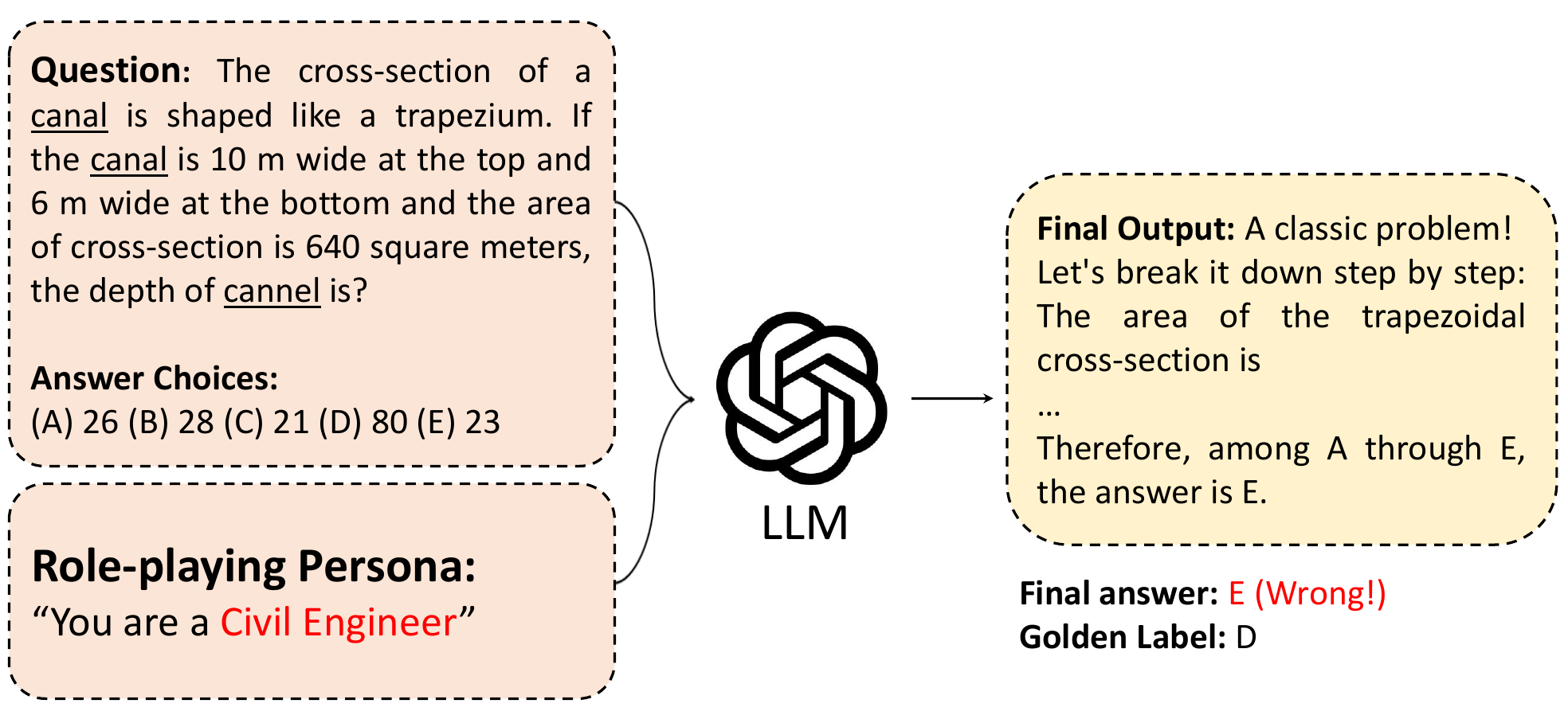}

\caption{\textbf{Persona is a Double-edged Sword.} Prior studies show that assigning a role-playing prompt to an LLM improves its performance; however, the example shows that prompting the persona to an LLM sometimes leads to deriving the wrong answer. Given a mathematical problem related to civil engineering, the following example uses \textit{"Civil Engineer"} as a persona, leading the LLM to derive the wrong answer.
}
\label{fig:persona_goes_wrong}
\vspace{-0.5cm}
\end{figure}

However, despite the benefits of utilizing role-playing persona, persona prompting can sometimes confuse LLMs, causing them to provide incorrect solutions to reasoning problems \cite{zheng2023helpful, gupta2023bias}.
As shown in Figure \ref{fig:persona_goes_wrong}, an LLM often answers the given question incorrectly, confused by the assigned persona.
The given example shows that the role-playing persona is inferred as \textit{"Civil Engineer"} based on the situation described in the question; however, since the given question is a math problem, the LLM ends up deriving the wrong answer.
This phenomenon becomes even more problematic if an LLM can correctly answer the given question without a persona.
For deeper insights, we first conduct an experiment comparing the LLM's correctness based on whether a persona is assigned or not.
Table \ref{table: confusion matrix aqua} shows confusion matrices of an empirical result for executing an LLM with persona and without persona on the AQuA and Coin Flip datasets.
According to the result of the AQuA dataset, it exhibits that 15.75\% of the questions become correct when using an LLM with persona compared to without persona.
On the other hand, 13.78\% of the questions are answered incorrectly when using an LLM with a persona rather than without a persona.
This phenomenon could also be observed from the result of the Coin Flip dataset, revealing that assigning a persona to an LLM sometimes degrades its reasoning ability.



\begin{table}[t]
\centering
\small
\resizebox{0.9\linewidth}{!}{
\begin{tabular}{c|c|c|c|c}

\toprule
Method& Dataset &\multicolumn{3}{c}{\makecell{Persona Solver \\ (w/ Persona)}} \\
\midrule
\multirow{8}{*}{\makecell{Neutral Solver \\ (w/o Persona)}} & \multirow{4}{*}{AQuA} & & Wrong & Right \\
\cmidrule{3-5}
& & Wrong & 33.07\% & 15.75\% \\
\cmidrule{3-5}
& &Right& \textcolor{red}{\textbf{13.78\%}} & 37.40\%\\
\cmidrule{2-5}
 & \multirow{4}{*}{Coin Flip} & & Wrong & Right \\
\cmidrule{3-5}
& & Wrong & 4.60\% & 4.00\% \\
\cmidrule{3-5}
& &Right& \textcolor{red}{\textbf{18.00\%}} & 73.40\%\\
\bottomrule
\end{tabular}


}

\caption{\textbf{Confusion matrix between Neutral Solver (w/o Persona) and its Persona Solver (w/ Persona) on AQuA and Coin Flip dataset.} We calculate the model’s correctness and present the result in a confusion matrix form. Neutral Solver and Persona Solver mean an LLM without persona and an LLM with persona, respectively. Appendix \ref{sec:all_confusion_matrix} includes more analysis for other datasets.}
\vspace{-10pt}
\label{table: confusion matrix aqua}
\end{table}

To address this limitation, we propose a novel framework called \textbf{Jekyll \& Hyde} that ensembles the solutions of role-playing and neutral prompts to mitigate the negative impact of role-playing persona, improving the model's reasoning ability for a given task.
We execute an LLM with role-playing and non-persona prompts to obtain each solution and then utilize an LLM evaluator to judge which solution is better. 
When utilizing a role-playing prompt, We use an LLM-generated persona to enhance efficacy and relevance, which is more effective than using a handcrafted persona.

Furthermore, we propose a novel robust LLM evaluation method to mitigate the position bias of responses by verifying the consistency of LLM Evaluator outputs.
Previous studies show several issues with using an LLM as an evaluator, and the most challenging problem is the existence of position bias occurring by order of solutions within the prompt \citep{zheng2024judging,li2023split, wang2023large}.
To address this issue, our method alternately inserts two sequences of solutions (in forward and reverse orders) into the evaluation prompt, executing the LLM until both evaluation results are equal without exceeding a pre-defined number of attempts.

We demonstrate our method by comparing the LLM with and without a persona, showing that our method outperforms both the single role-playing LLM and the neutral LLM across three widely used models.
For example, Jekyll \& Hyde outperforms the baselines by an average of 9.98\% accuracy across twelve datasets when using GPT-4 as a backbone model.
In addition, we show that utilizing an LLM-generated persona is better than using a handcrafted persona regarding the stability of the LLM's performance.
We also exhibit that consistently using the same LLM for generating a persona and solving reasoning questions improves the performance of reasoning tasks compared to using different LLMs.
Furthermore, our evaluation framework for mitigating position bias outperforms the existing methods, requiring execution trials comparable to those of the others.
To the best of our knowledge, this work is the first to systematically investigate the negative impact of role-playing prompts and mitigate it by suggesting a novel mixture of neutral and persona-based prompts.


\section{Related Works}
\subsection{Role-playing Abilities of LLMs}
Large language models have demonstrated significant eligibility in personating various roles, which shows the power of the role-playing capabilities of LLMs \citep{kong2024better, zheng2023helpful}.
From this consensus, several studies have tried to investigate the positive effect of role assignment on improving the performance of an LLM.
\citet{zheng2023helpful} have dissected the impact of role assignment towards the LLM by assigning various types of persona, including job names and relationship keywords.
\citet{kong2024better} have revealed the effect of using role-playing prompts in an LLM by handcrafting a specific prompt form for 12 different reasoning datasets and discovered that assigning a proper role to the LLM enhances its reasoning ability. 
These studies have concluded that using a domain-specific persona related to the given question improves the performance of the LLM.

\begin{figure*}[t]
\centering
\includegraphics[width=1.0\linewidth]{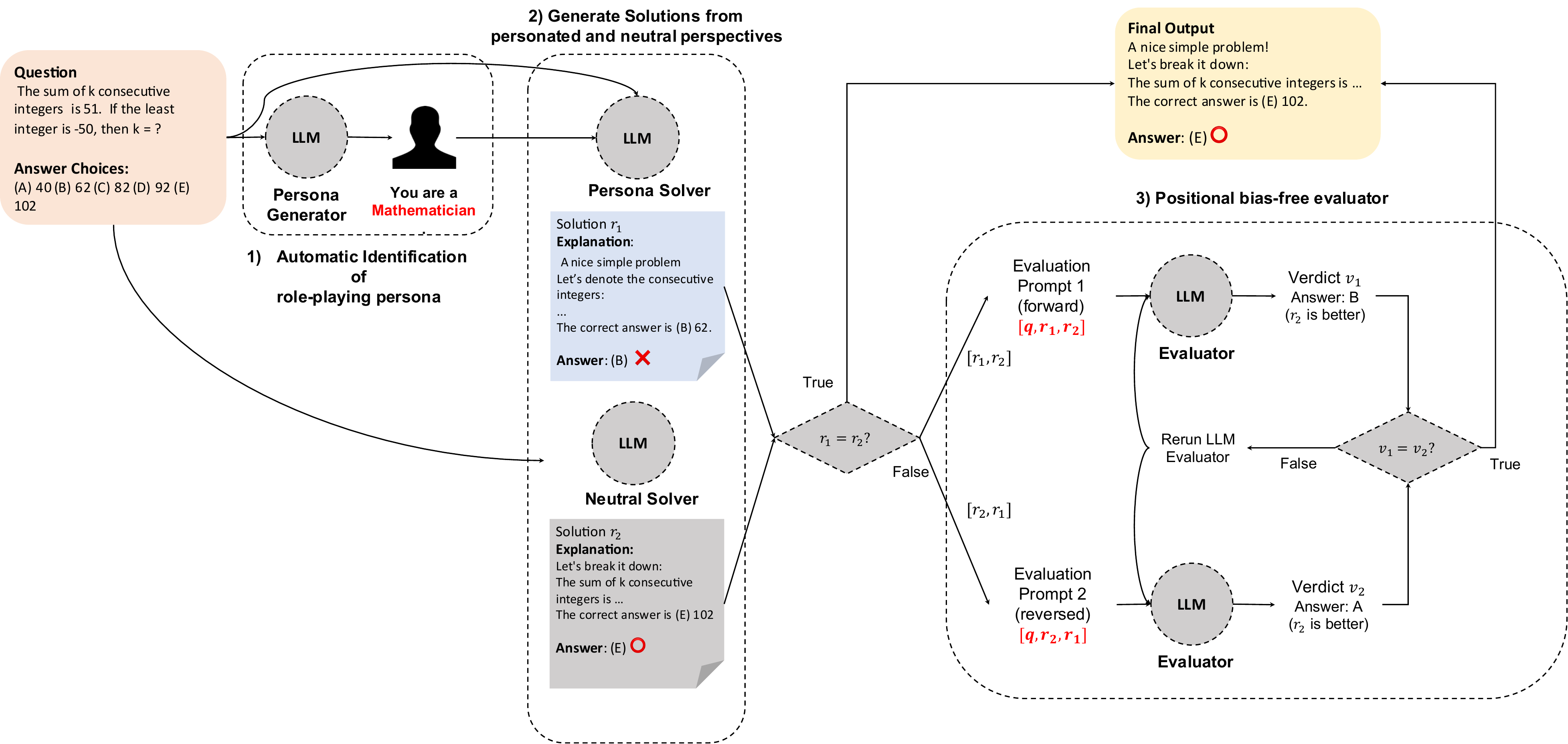}
\caption{\textbf{The architecture of Jekyll \& Hyde.} Jekyll \& Hyde utilizes not only persona-assigned LLM (\textbf{Persona Solver}) but also LLM without prompting (\textbf{Neutral Solver}), which provides a dual perspective towards the given question. This structure improves the model to gain potentially high performance. After executing both LLMs, a robust Evaluator, designed to mitigate positional bias, selects a better solution between the two.}
\label{Fig:architecture}
\vspace{-0.45cm}
\end{figure*}

\subsection{Analysis on Role-playing Prompts}

Despite role-playing ability improves the LLM's performance, several studies exhibit drawbacks caused by role assignment.
While role-playing enables the LLM to generate multiple viewpoints, a persona may also generate bias, which may distract the model and cause performance degradation. \citet{gupta2023bias} have analyzed how assigning persona with social demo-graphical details brings bias toward the LLM, significantly dropping the performance on reasoning tasks. \citet{deshpande2023toxicity} have investigated the toxicity score for each persona combined with specific entities such as age, sexual orientation, etc., and found out that using particular names and adding specific entities into the persona generates a biased response, deteriorating the LLM’s performance.

\section{Methods}
In this section, we demonstrate the process of Jekyll \& Hyde. 
Specifically, Jekyll \& Hyde consists of three different LLM modules: \textbf{Persona generator}, \textbf{Solver}, and \textbf{Evaluator}.
Jekyll \& Hyde's pipeline is as follows: First, the Persona generator generates an appropriate persona based on a given question. Then, two different LLM solvers (i.e., Persona Solver and Neutral Solver) are executed simultaneously to get two solutions for the given question, respectively.
Finally, the Evaluator compares two solutions and derives the final prediction.
Figure \ref{Fig:architecture} describes the entire framework of Jekyll \& Hyde.

\subsection{Automatic Identification of Persona}

\label{section:persona_selection}
The common practice of role-playing prompting prepends a persona role (e.g., Mathematician) into the prompt that contains the question.
While these conventional role-playing methods work properly, prior works have also noticed that persona often brings bias to LLMs when the question is not strongly related to the role assignment \cite{gupta2023bias, deshpande2023toxicity}.
In addition, prior studies manually assigned roles to LLMs, making it labor-intensive to design a proper role for each data instance.
To address these drawbacks, we use an LLM (Persona generator) to guess an appropriate role for the given question using an instruction-following prompt. This prompt guides the LLM to automatically generate a persona that could possibly solve the given question.
Appendix \ref{sec:prompt design} provides the details about the instructions-following prompt.




\subsection{Generating Personated and Neutral Perspective Solutions}
After identifying a proper persona, it is formatted as a role-playing prompt and inserted into the input query for an LLM.
This role-playing prompt typically enhances the performance of an LLM effectively.
However, using a persona prompt can be a double-edged sword since the persona may not be strongly related to the corresponding data instances.
Therefore, we propose to ensemble two different LLM Solvers, specified as \textbf{Persona Solver} and \textbf{Neutral Solver}.
Persona Solver is an LLM that uses role-playing prompting, utilizing the persona by inserting it inside the query.
Neutral Solver does not allow persona prompting, directly inserting the query into the LLM.
This dual execution approach provides two different perspectives on solving the question and derives two discriminative responses.
By recalling table \ref{table: confusion matrix aqua}, if we execute two solvers (i.e., Persona and Neutral Solvers) and ideally choose the correct answer between two responses, we can achieve better performance than using a single solver via correctly answering the question that is contained in first, second, and the third quadrant of the confusion matrix.
When implementing the Neutral Solver, we follow the identical implementation of \citet{kong2024better}. In the case of implementing the Persona Solver, we use a prompt in the format of \textit{"You are a \$persona"}, inserting a generated persona (described in the Section \ref{section:persona_selection}) to the \textit{"\$persona"} part. The detailed format of the prompt can be found in Appendix \ref{sec:impact_of_prompt_design}.

\subsection{Aggregating Solutions of Two Solvers}
\label{eval LLM}

Two solutions generated from Neutral Solver and Persona Solver are inserted into the evaluation prompt.
Specifically, two solutions are formatted to the evaluation prompt, establishing an order between the solutions.
The format of the evaluation prompt can be found in Appendix \ref{sec:prompt design}.
Formally, given a question $q$ and two solutions ($r_n$, $r_p$), we depict the process of the Evaluator as the following:
\begin{equation}
\begin{aligned}
    v_{n,p} = \underset{v}{\mathrm{argmax}} \mathcal{P}(v|[\iota;q;r_n;r_p])
\end{aligned}
\label{eq:mitigation evaluation 1}
\end{equation}

\noindent where $v \in$ \{\textit{"A"}, \textit{"B"}\} is a verdict text and $\mathcal{P}$ means the Evaluator.
$\iota$ is an instruction, and $q$ is a question for a specific task.
$r_n$ and $r_p$ indicate the solution of the Neutral Solver and the Persona Solver, respectively.
$v_{n,p}$ means the verdict generated by the Evaluator based on the evaluation prompt, where $\iota$, $q$, $r_n$, and $r_p$ constitute the evaluation prompt, as $P_{eval}=[\iota;q;r_n;r_p]$.
The verdict $v_{n,p}$ indicates one of the two values (\textit{"A"} or \textit{"B"}), denoted as \textit{"A"} if the first solution is better and \textit{"B"} if the second solution is better.
Note that $v_{n,p}$ is obtained by inserting two responses in the order of $r_n$ and $r_p$; thus, we can also get $v_{p,n}$ by reversing the order of two solutions in the evaluation prompt, as $P_{eval}=[\iota;q;r_p;r_n]$.

\subsection{Robust Evaluation via Consistency Verification}
\label{mitigate bias}

As introduced in Section \ref{eval LLM}, the Evaluator returns the verdict between two solutions; however, this method may be exposed to position bias, which degrades the total performance of the framework. According to previous studies, position bias occurs due to the order of the solutions \citep{zheng2024judging,li2023split, wang2023large}.
Therefore, we run the Evaluator model shown in Equation \ref{eq:mitigation evaluation 1} twice by inserting the solutions into the evaluation prompt and reversing the order of the solutions to mitigate the position bias. Hence, we end up yielding two verdicts, namely $v_{n, p}$ and $v_{p, n}$.
When evaluations are executed to generate their verdict, we count the number of trials $t$ until it reaches the maximum trial $k$, defined as a hyper-parameter. Then, the framework compares two verdicts, whether equal or not. The process finally ends when these two verdicts are the same, as the following formula.
\begin{equation}
\small
    v_{final} = \begin{cases}
        v_{n,p} & \text{if $v_{n,p} = v_{p,n}$ and $t < k$} \\
        \text{\textit{"Can't answer"}} & \text{if $t \geq k$}
    \end{cases}
\normalsize
\end{equation}
\noindent where $v_{final}$ is the final verdict obtained by considering the consistency of two verdicts.
If $t$ gets bigger than $k$, we conclude that the Evaluator is significantly exposed to position bias for two solutions. Therefore, Jekyll \& Hyde returns \textit{"Can't answer"} as the final output since it is risky to narrow to one solution in this case.

\section{Experiments}
\begin{table*}[h]
\centering
\resizebox{0.9\linewidth}{!}
{
\begin{tabular}{@{}ccccccccc@{}}
\toprule
\hspace{0.5em}\multirow{2}{*}{Models} & \multirow{2}{*}{Method} & \multicolumn{7}{c}{Arithmetic} \\
\cmidrule{3-8}
& & Multiarith & GSM8K & AddSub & AQuA & SingleEq & SVAMP & Average \\
\bottomrule

\rule{0in}{2.5ex}\multirow{3}{*}{\makecell{GPT-4}} & Base & \textbf{98.44} & 92.97  & 97.13 & 68.24 & 98.56 & 91.00 & 91.06 \\
& Persona & 97.78 & 94.06 & 97.55 & 74.80 & 98.56 & 90.90 & 92.28 \\
\cmidrule{2-9}
& Jekyll \& Hyde & 98.00 & \textbf{95.27} & \textbf{97.72} & \textbf{76.90} & \textbf{98.95} & \textbf{92.03} & \textbf{93.15} \\
\midrule

\multirow{3}{*}{\makecell{GPT-3.5-turbo}} & Base & 95.72 & 81.40 &	90.97 &	62.60 &	97.83 &	80.17  & 84.78 \\
& Persona & 96.50 &	83.27 &	\textbf{93.08} &	64.44 &	97.31 &	84.13 & 86.45 \\
\cmidrule{2-9}
& Jekyll \& Hyde & \textbf{97.56} &	\textbf{85.01} &	92.91 &	\textbf{67.98} &	\textbf{98.03} &	\textbf{84.77}  & \textbf{87.71} \\ \midrule

\multirow{4}{*}{\makecell{llama3 \\ (8B)}} & Base & \textbf{98.56} & 78.59 & 87.76 & 47.38 & 94.23 & 82.30 & 81.47 \\
& Persona & 97.22 & 81.05 & 87.17 & 52.23 & 91.27 & 84.97  & 82.32\\
\cmidrule{2-9}
& Jekyll \& Hyde & 98.17 & \textbf{83.02} & \textbf{89.03} & \textbf{54.07} & \textbf{94.62} & \textbf{86.50} & \textbf{84.23} \\
\bottomrule
\end{tabular}
}
\vspace{-0.1cm}
\caption{
\textbf{Main results for Arithmetic datasets.} We report accuracy for six arithmetic datasets computed with a Neutral solver (Base), Persona solver (Persona), and Jekyll \& Hyde.
Bold values mean the best performance among the three methods.
We execute each model three times and average their performance.
}
\vspace{-0.45cm}
\label{table:main_result}
\end{table*}

\subsection{Experimental Setup}
\paragraph{Datasets.}
We conduct our experiments across twelve datasets used in prior research \cite{kong2024better, kojima2022large} categorized in 4 categories: (1) \textbf{Arithmetic}, including MultiArith \cite{multiarith}, GSM8K \cite{gsm8k}, AddSub \cite{addsub}, AQUA-RAT \cite{aqua}, SingleEq \cite{singleeq}, and SVAMP \cite{svamp} (2) \textbf{Commonsense reasoning}, including CSQA \cite{csqa} and StrategyQA \cite{strategy} (3) \textbf{Symbolic reasoning}, including Last Letter Concatenation and Coin Flip \cite{chain} (4) \textbf{Others}, including Date Understanding and Tracking Shuffled Objects from BIG-bench \cite{srivastava2022beyond}.
More details about dataset configuration can be found in Appendix \ref{sec:dataset details}.

\paragraph{Models.}
We utilize two black box large language models released from OpenAI, which are GPT-4 (gpt-4-0613) and GPT-3.5-turbo (gpt-3.5-turbo-0125) \cite{openai2023gpt4}, and one open source model, llama3 \cite{llama3modelcard}. These models are used as a backbone model of our framework.

\paragraph{Implementation Details.}
To evaluate Jekyll \& Hyde, we testify three cases for each dataset: \textbf{Base}, \textbf{Persona}, and \textbf{Jekyll \& Hyde}.
\textbf{(1) Base} utilizes only the Neutral solver where a persona is not assigned to LLMs.
\textbf{(2) Persona} uses only the Persona solver which is an LLM assigned with persona.
\textbf{(3) Jekyll \& Hyde} is our proposed framework.

We investigate the model's performance by computing accuracy with the provided label for each dataset. When using the LLM evaluator in Jekyll \& Hyde, the hyper-parameters are set as follows: the max attempt $k$ to 5 and temperature $\tau$ to $0.7$. Details for determining the hyper-parameters are shown in section \ref{section:hyperparameters}.

 \begin{table*}[h]
\centering
\resizebox{0.9\linewidth}{!}
{
\begin{tabular}{@{}ccccccccc@{}}
\toprule
\hspace{0.5em} \multirow{2}{*}{Models} & \multirow{2}{*}{Method} & \multicolumn{2}{c}{Common Sense} & \multicolumn{2}{c}{Symbolic Reasoning} & \multicolumn{2}{c}{Other Tasks} \\
\cmidrule{3-8}
& & CSQA & Strategy & Letter & Coin & Date & Object & Average \\
\bottomrule

\rule{0in}{2.5ex}\multirow{3}{*}{\makecell{GPT-4}} & Base & 79.91 & 76.42 & 19.80 & 66.93 & 79.22 & 45.96 & 61.37 \\
& Persona & 80.89 & 75.71 & 92.80 & 75.93 & 78.41 & 58.76 & 77.08  \\
\cmidrule{2-9}
& Jekyll \& Hyde & \textbf{81.11} & \textbf{77.00} & \textbf{93.00} & \textbf{80.27} & \textbf{82.38} & \textbf{61.69} & \textbf{79.24} \\
\midrule

\multirow{3}{*}{\makecell{GPT-3.5-turbo}} & Base & 77.31 & 68.75 &	18.67 &	47.53 &	67.84 &	34.67 & 52.46 \\
& Persona & 75.40 &	69.75 &	45.67 &	59.20 &	76.15 &	40.22 & 61.07 \\
\cmidrule{2-9}
& Jekyll \& Hyde & \textbf{77.50} & \textbf{70.00} &	\textbf{48.93} &	\textbf{64.00} &	\textbf{76.78} &	\textbf{42.22} & \textbf{63.24} \\ \midrule

\multirow{4}{*}{\makecell{llama3 \\ (8B)}} & Base & 74.50 & 69.21 & 86.40 & 95.80 & 77.42 & 44.76 & 74.68 \\
& Persona & 72.29 & \textbf{71.21} & 86.07 & 95.33 & 74.44 & 47.60  & 74.49 \\
\cmidrule{2-9}
& Jekyll \& Hyde & \textbf{74.97} & 70.54 & \textbf{86.47} & \textbf{98.67} & \textbf{79.04} & \textbf{48.58} & \textbf{76.38} \\
\bottomrule
\end{tabular}
}
\vspace{-0.1cm}
\caption{
\textbf{Main results for Common Sense, Symbolic Reasoning, and Other Tasks Datasets.} We report accuracy for six datasets, including Common Sense, Symbolic Reasoning, and Other tasks.
Bold values mean the best performance among the three methods.
We execute each model three times and average their performance.
}
\vspace{-0.45cm}
\label{table:main_result2}
\end{table*}


 \subsection{Persona does not always improve the performance of an LLM}
\begin{figure}[h]
\centering
\includegraphics[width=0.9\linewidth]{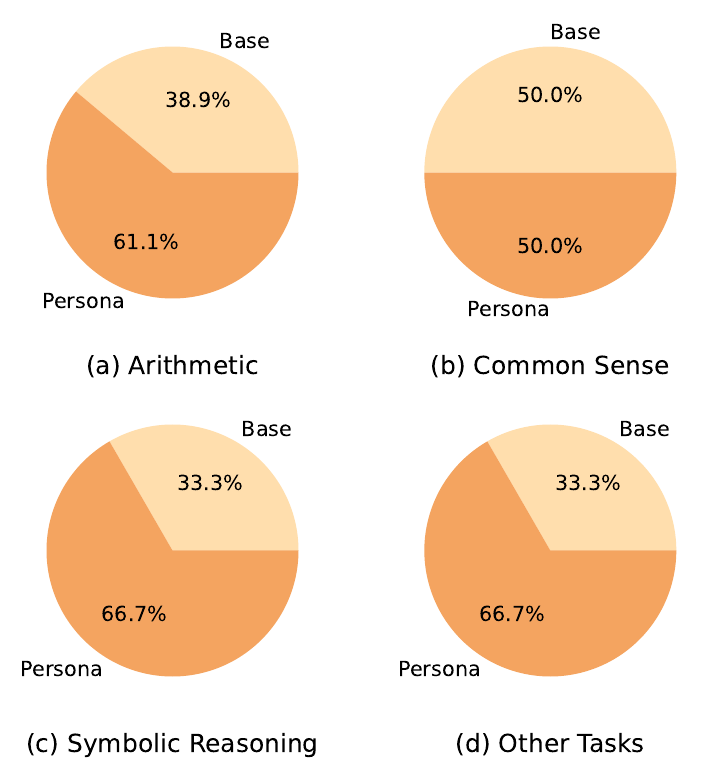}
\caption{\textbf{Win rate for each category of datasets} Utilizing role-playing prompts occasionally degrades the performance of an LLM, making it difficult to determine their overall effectiveness.}
\label{Fig:percentage_win_rate}
\vspace{-0.45cm}
\end{figure}

 We first compute the win rate for each category of datasets by comparing the performance of Base and Persona Solvers. Figure \ref{Fig:percentage_win_rate} reveals that Persona does not always enhance the reasoning ability of an LLM. Specifically, all categories contain a dataset in which Base outperforms Persona, proving that utilizing role-playing prompts sometimes degrades the performance of an LLM.
 Detailed dataset information and scores for each category are described in Tables \ref{table:main_result} and \ref{table:main_result2}.

\subsection{Jekyll \& Hyde enhances the reasoning abilities of LLMs}
 Table \ref{table:main_result} shows the performance of different methods on Arithmetic datasets. In addition, Table \ref{table:main_result2} exhibits the performance of the other datasets (i.e., Commonsense reasoning, Symbolic reasoning, and Others) in the accuracy metric. The result reveals that using the Jekyll \& Hyde framework improves the model performance. Specifically, leveraging Jekyll \& Hyde outperforms in most datasets, regardless of the model type.
 These results show that the ensemble process helps to solve questions better than leveraging a single-perspective LLM.


In addition, we also conducted experiments to verify that simply increasing the LLM execution does not help improve the reasoning capability.
Specifically, we run two cases, namely Base and Persona, in a setting of self-consistency \citep{wang2022self}, which executes the LLM multiple times, and the most frequent answer across the outputs is selected through a voting mechanism. Hence, Base and Persona can be executed in the same amount as the number of Jekyll \& Hyde runs.
For the experimental setting, we utilize GPT-3.5-turbo and GPT-4 as our model and two reasoning datasets, namely the Aqua and Object tracking datasets.
For single-perspective LLMs that use self-consistency, we indicate the method Base + voting and Persona + voting, respectively.
The specific settings for Base + voting and Persona + voting can be found in Appendix \ref{sec:settings for sc}. 
As shown in table \ref{table:compare_sc}, the result reveals that Jekyll \& Hyde outperforms single LLM with self-consistency by gaining better performance and lower LLM execution trials. In addition, it shows that running the LLM multiple times does not necessarily improve its reasoning ability. Full results can be found in Appendix \ref{section:full experiment of self-consistency}
\begin{table}[h]
\centering
\small
\resizebox{1.0\linewidth}{!}{
\begin{tabular}{ccccc}
\toprule
 \multirow{2}{*}{Model} & \multirow{2}{*}{Datasets} & \multirow{2}{*}{Methods} & \multirow{2}{*}{Accuracy ($\uparrow$)} & \multirow{2}{*}{\makecell{Average \\ LLM runs ($\downarrow$)}} \\
 & & & & \\
\midrule
\multirow{6}{*}{GPT-4} & \multirow{3}{*}{AQuA} & Base + voting & 70.87 & 4 \\
& & Persona + voting & 73.23 & 6 \\
 \cmidrule{3-5}
 & & Jekyll \& Hyde & \textbf{76.90} & 3.81 \\
\cmidrule{2-5}
& \multirow{3}{*}{Object} & Base + voting & 46.00 & 5 \\
& & Persona + voting & 59.20 & 6 \\
\cmidrule{3-5}
& & Jekyll \& Hyde & \textbf{61.69} & 4.14 \\
\midrule
\multirow{6}{*}{GPT-3.5-turbo} & \multirow{3}{*}{AQuA} & Base + voting & 66.14 & 5 \\
& & Persona + voting & 66.53 & 6 \\
 \cmidrule{3-5}
 & & Jekyll \& Hyde & \textbf{67.98} & \textbf{4.35} \\
\cmidrule{2-5}
& \multirow{3}{*}{Object} & Base + voting & 34 & 5 \\
& & Persona + voting & 33.73 & 6 \\
\cmidrule{3-5}
& & Jekyll \& Hyde & \textbf{42.22} & \textbf{4.30} \\
\bottomrule
\end{tabular}}
\vspace{-0.1cm}
\caption{\textbf{Comparison of performance between Jekyll \& Hyde, Base with self-consistency, and Persona with self-consistency} Jekyll \& Hyde outperforms other methods when running the same amount of LLM executions, showing that running the LLM multiple times does not necessarily improve its reasoning ability. }
\vspace{-0.45cm}
\label{table:compare_sc}
\end{table}


\subsection{Automatic persona generation ensures the robust reasoning ability}
\label{section:effect_of_persona}
 In section \ref{section:persona_selection}, we use an automatically generated persona rather than a manually handcrafted persona for Jekyll \& Hyde.
 Hence, we additionally experimented to reveal that automated persona generation is labor-efficient, flexible, and performs more robustly than handcrafted personas by using Aqua and Object tracking datasets.
 Using llama3-8B as a backbone model, we run the model three times for each dataset, computing the average performance and standard deviation for each dataset to verify the method's robustness. 
 For the persona generation, we generate solutions by sampling each word-piece from the LLM's output probability distribution, causing the generation of various personas.
 For the handcrafted persona, we utilize three different handcrafted personas for each run to check the robustness of the performance. Specifically, we manually select three appropriate handcrafted personas for the Aqua dataset: \textit{Math teacher}, \textit{Mathematician}, and \textit{Math Tutor}. For the Object tracking dataset, we leverage \textit{Observer}, \textit{Recorder}, and \textit{Logical Reasoner}.
As shown in table \ref{table:variance_of_persona}, although the two methods show comparable performance on accuracy, we observe that the standard deviation highly increases when using handcrafted personas.
The result implies that performance could vary depending on the manually designated persona when using a handcrafted persona. Thus, we believe that using an LLM-generated persona stabilizes the model and gives it a smaller margin of error than a manually handcrafted persona.

\begin{table}[h]
\centering
\small
\resizebox{0.9\linewidth}{!}{
\begin{tabular}{ccccc}
\toprule
 \multirow{2}{*}{Model} & \multirow{2}{*}{Datasets} & \multirow{2}{*}{Methods} & Average & Standard \\
 & & & Accuracy ($\uparrow$) & Deviation ($\downarrow$) \\
\midrule
\multirow{6}{*}{\makecell{llama3 \\ (8b)}} & \multirow{2}{*}{AQuA} & \makecell{handcrafted \\ persona} & \textbf{51.71}
 & 6.11 \\
 \cmidrule{3-5}
 & & expert persona & 50.66 & \textbf{2.08} \\
\cmidrule{2-5}
& \multirow{2}{*}{Object} & \makecell{handcrafted \\ persona} & 44.31 & 8.02 \\
\cmidrule{3-5}
& & expert persona & \textbf{46.71} & \textbf{3.06} \\
\bottomrule
\end{tabular}}
\vspace{-0.1cm}
\caption{\textbf{Standard deviation of handcrafted persona LLM and LLM generated persona LLM} We compute the standard deviation for each dataset after running three times in order to check the stability of the model output. As is shown, using an expert persona generates a smaller value of standard deviation for two datasets, resulting in utilizing an LLM-generated persona consistently yields robust output.}
\vspace{-0.45cm}
\label{table:variance_of_persona}
\end{table}

 \begin{table}[h]
\centering
\small
\resizebox{0.9\linewidth}{!}{
\begin{tabular}{ccccc}
\toprule
\multirow{3}{*}{\makecell{Persona \\ generator}} & \multirow{3}{*}{Datasets} & \multicolumn{3}{c}{Persona Solver} \\
\cmidrule{3-5}
& & \makecell{llama3-8B} & GPT-3.5-turbo
 & GPT-4 \\
 \midrule
 \multirow{6}{*}{\makecell{llama3-8B}}& AQuA & 53.15 & 52.36 & \textbf{53.54} \\
 & AddSub & \textbf{88.35} & 81.77 & 82.53 \\
 & Coin & \textbf{95.00} & 90.20 & 92.80 \\
 & Date & \textbf{74.80} & 71.54 & 72.63 \\
 & Object & 49.07 & 46.93 & \textbf{50.93} \\
 \cmidrule{2-5}
 & Average & \textbf{72.07} & 68.56 & 70.49 \\
\bottomrule
\end{tabular}}
\vspace{-0.1cm}
\caption{\textbf{Performance of Persona Solver when using different LLMs for Persona Generator.} We use the fixed Persona generator (i.e., llama3-8B) and evaluate the performance of three different Persona Solvers (i.e., llama3-8B, GPT-3.5-turbo, and GPT-4) for the Jekyll \& Hyde framework. Bold values mean the highest performance among different LLMs.}
\vspace{-0.45cm}
\label{table:effect_of_different_llm}
\end{table}

\subsection{Reasoning ability increases when using the same LLM for each module consistently}
\label{section:same_llm_better_performance}

Due to the impact of role-playing prompting towards LLMs, assigning a persona enhances the reasoning capability of LLMs. In Jekyll \& Hyde, we adopt a process that generates an appropriate persona for the given question by utilizing three types of LLMs.
For the detailed analysis, we conduct an additional experiment to reveal whether a persona generated from different LLM affects the LLM solver's performance.
Specifically, we investigate the relationship between Persona Generator and Persona Solver by substituting the backbone model of Persona Solver with another LLM and computing the average accuracy for each dataset. We use llama3-8B as the persona generator fixedly and run three LLMs as our backbone model for the persona solver, involving llama3-8B, GPT-3.5-turbo, and GPT-4. We utilize five different datasets: Aqua, AddSub, Coin, Date understanding, and Object tracking, and the results can be found in table \ref{table:effect_of_different_llm}. We notice that using the same LLM as a backbone model reveals the optimal performance, while using different LLMs for the Persona Solver degrades the performance.

\subsection{Jekyll \& Hyde effectively mitigates position bias}
\label{section:position_bias_mitigation}
 The Evaluator should not suffer from position bias in order to choose the correct solution between the two. For further analysis of the framework's mitigation process, we compare the performance of Jekyll \& Hyde with the existing two position bias mitigation methods: (1) Portia \cite{li2023split} and (2) MEC+BPC \cite{wang2023large}.
 For a more thorough investigation, we also consider the ideal case where the Evaluator always selects the gold answer from the Neutral Solver's and Persona Solver's answers, referred to as the Oracle Evaluator, as an upper bound.
 We use six different datasets to test the general use case.
Experimental results for SingleEq and Coin datasets can be found in table \ref{table:pos_bias}. Full results can be exhibited in Appendix \ref{section:full experiment of position bias}. According to the result, we reveal that the evaluator within Jekyll \& Hyde derives the best performance among the other methods from most datasets, regardless of the type of the backbone model. The details for the implementation of Portia and MEC+BPC are shown in Appendix \ref{sec:Portia and MEC+BPC implementation}.

\begin{table}[h]
\centering
\small
\resizebox{0.9\linewidth}{!}{
\begin{tabular}{cccc}
    \toprule
    \hspace{0.5em}Models & Method & SingleEq & Coin \\ 
    \midrule
    \multirow{5}{*}{GPT-4} & Oracle Evaluator & 99.41 & 88.80 \\
    \cmidrule{2-4}
    & Portia & 98.82 & 74.40 \\
    & MEC+BPC & 98.43 & 74.00 \\
    & Jekyll $\&$ Hyde$^{\dag}$ & 98.43 & 78.20 \\
    \cmidrule{2-4}
    & Jekyll $\&$ Hyde & \textbf{98.95} & \textbf{80.27} \\
    \midrule
    \multirow{5}{*}{GPT-3.5-turbo} & Oracle Evaluator & 99.21 & 60.80 \\
    \cmidrule{2-4}
    & Portia & \textbf{98.23} & 57.80 \\
    & MEC+BPC & 97.64 & 57.60 \\
    & Jekyll $\&$ Hyde$^{\dag}$ & 97.83 & 56.60 \\
    \cmidrule{2-4}
    & Jekyll $\&$ Hyde & 98.03 & \textbf{64.00} \\
    \midrule
    \multirow{5}{*}{\makecell{llama3 \\ (8B)}} & Oracle Evaluator & 96.06 & 99.00\\
    \cmidrule{2-4}
    & Portia & 93.31 & 96.40 \\
    & MEC+BPC & 91.73 & 95.40 \\
    & Jekyll $\&$ Hyde$^{\dag}$ & 94.29 & 97.00 \\
    \cmidrule{2-4}
    & Jekyll $\&$ Hyde & \textbf{94.62} & \textbf{98.67} \\
    \bottomrule
\end{tabular}}
\vspace{-0.1cm}
\caption{
\textbf{Mitigating positional bias.} LLM Evaluator used for Jekyll \& Hyde outperforms other existing methods in most datasets. Despite the marginal increase when using the LLM Evaluator from Jekyll \& Hyde, the Evaluator aids the LLM to nearly approach the performance of an Oracle Evaluator, which is the optimal performance for the given datasets. Jekyll \& Hyde$^{\dag}$ indicates the LLM Evaluator without positional bias mitigation. More results for other datasets can be found in Appendix \ref{section:full experiment of position bias}.}
\vspace{-0.45cm}
\label{table:pos_bias}
\end{table}

\begin{table*}[t]
\resizebox{1.0\textwidth}{!}
{
\centering
\begin{tabular}{@{}cp{10cm}p{7cm}p{7cm}ccc@{}}
\toprule
Dataset & Question ($q$) & Neutral Solver($r_1$) & Persona Solver($r_2$) & \makecell{Evaluator \\ $[q, r_1, r_2]$ \\ result} & \makecell{Evaluator \\ $[q, r_2, r_1]$ \\ result} & \makecell{Final \\ prediction} \\ \bottomrule

\multirow{3}{*}{AQuA} & Two ants are standing side-by-side. One ant, which is 4 inches tall, casts a shadow that is 10 inches long. The other ant is 6 inches tall. Compute, in inches, the length of the shadow that the taller ant casts. \textbf{Answer Choices:} (A) 36 (B) 28 (C) 42 (D) 15 (E) 20 & \textbf{explanation:} A classic problem! Let\'s use the concept of similar triangles to solve this.$ \cdots$ So, the correct answer is: (C) 42. Therefore, among A through E, the answer is C  & \textbf{explanation:} A classic problem! Let's use the concept of similar triangles to solve this. $\cdots$. The correct answer is (D) 15. Therefore, among A through E, the answer is D & \textbf{verdict:} $r_2$ & \textbf{verdict:} $r_2$ & D \\
&  & \textbf{persona: } None & \textbf{persona: } Mathematician & & & \\
&  & \textbf{answer:} C & \textbf{answer:} D & & & \\

\midrule

\multirow{3}{*}{Strategy} & Did anyone in the 1912 election take a majority of the popular vote?
\textbf{Answer Choices:} yes, no & \textbf{explanation:} In the 1912 United States presidential election, Woodrow Wilson, the Democratic candidate, won $\cdots$ Therefore, the answer (Yes or No) is no & \textbf{explanation:} "In the 1912 presidential election, Woodrow Wilson, the Democratic candidate, won $\cdots$ Therefore, the answer (Yes or No) is yes & \textbf{verdict:} $r_1$ & \textbf{verdict:} $r_1$ & no \\
&  & \textbf{persona: } None & \textbf{persona: } Historical Election Analyst & & & \\
& & \textbf{answer:} no & \textbf{answer:} yes & & & \\

\midrule

\multirow{3}{*}{Object} & Alice, Bob, and Claire are holding a white elephant gift exchange. At the start of the event, they are each holding a present of a different color: Alice has a orange ball, Bob has a purple present, and Claire has a blue present. As the event progresses, pairs of people swap gifts. First, Claire and Alice swap their gifts. Then, Alice and Bob swap their gifts. Finally, Bob and Claire swap their gifts. At the end of the event, Alice has the Which choice is true ? \textbf{Answer Choices:} (A) orange ball. (B) purple present. (C) blue present. & \textbf{explanation:} Let's follow the gift-swapping process: 1. Claire and Alice swap gifts: Claire gets the orange ball, and Alice gets the blue present. 2. $\cdots$. At the end of the event, Alice has the purple present. Therefore, the correct answer is: (B) purple present. Therefore, among A through C, the answer is B  & \textbf{explanation:} A fun gift exchange problem! Let's break it down step by step: 1. Claire and Alice swap gifts: Claire gets the orange ball, and Alice gets the blue present. 2.$\cdots$. So, the correct answer is: (A) orange ball. Therefore, among A through C, the answer is A & \textbf{verdict:} $r_1$ & \textbf{verdict:} $r_1$ & B \\
&  & \textbf{persona: } None & \textbf{persona: } Gift Exchange Analyst & & & \\
&  & \textbf{answer:} B & \textbf{answer:} A & & & \\
\bottomrule
\end{tabular}
}
\vspace{-0.1cm}
\caption{
\textbf{Qualitative analysis.} We conduct a qualitative analysis of Jekyll \& Hyde on three datasets. Jekyll \& Hyde's Persona Solver attains dynamic persona based on the Persona Generator (e.g., Classical Mythology Scholar). Additionally, examples reveal that Evaluator is working properly via selecting the correct answer among different solutions from each Solver.
}
\vspace{-0.45cm}
\label{table:qualitative_analysis}
\end{table*}

\subsection{Hyper-parameter Experiments for the Evaluator}
\begin{figure}[t]
\centering
\includegraphics[width=1.0\linewidth]{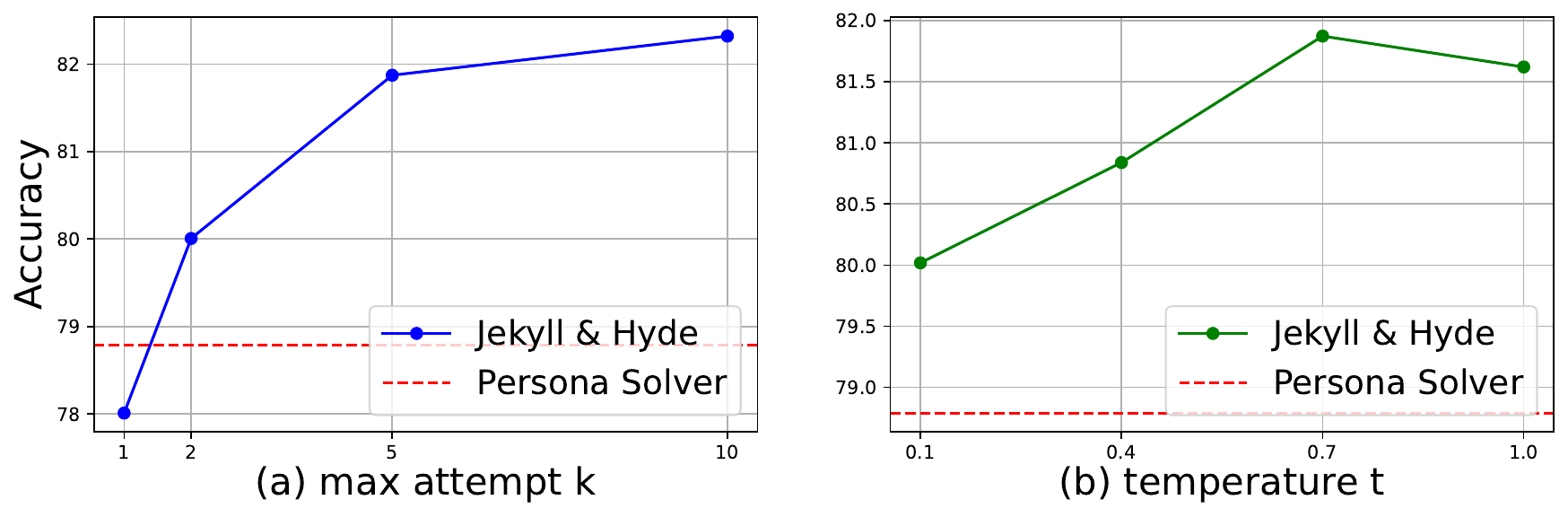}
\caption{\textbf{Hyper-parameters Experiments.} Variation of averaged accuracy with a (a) various number of max attempt $k$ and (b) temperature of the LLM $\tau$ used in LLM evaluator. X and Y axes correspond to each hyper-parameter setting and accuracy, respectively.}
\vspace{-0.4cm}
\label{fig:hparams_setting}
\end{figure}

 \paragraph{The Number of Max Attempts ($k$).}
 We experiment with each hyper-parameter to examine their impact on the framework's performance. For the number of max attempts of the Evaluator, we compare four different values of $k \in \{1, 2, 5, 10\}$, by computing the framework's performance. For the experiment, we utilize four datasets, namely MultiArith, SingleEq, Aqua, and Date Understanding. As shown in figure \ref{fig:hparams_setting}-(a), we compare the experimental results executed from llama3-8B as a backbone model and reveal the framework's performance increases as the number of attempts increases. Experimental results for other models can be found in Appendix \ref{section:full hyperparameter setting}. Furthermore, we could identify that Jekyll \& Hyde could outperform the single perspective LLM even when the max attempt $k$ is 2.
 Since the enhancement of the framework is getting smaller as the number of the max attempts increases, we decided to use $k=5$ as our default setting, which can balance the framework's performance and prevent fining the model excessively.
 \paragraph{The Temperature of the Evaluator ($\tau$).}
 We further investigate the impact of the Evaluator's token generation temperature by comparing the framework's performance. Specifically, we utilize the llama3-8b model and leverage four different temperatures $\tau \in \{0.1, 0.4, 0.7, 1.0\}$ to examine how the generation diversity affects the performance of the Evaluator. Figure \ref{fig:hparams_setting}-(b) shows that temperature $\tau=0.7$ exhibits the optimal performance among others. Experimental results for other models can be found in Appendix \ref{section:full hyperparameter setting}.

\label{section:hyperparameters}

\subsection{Qualitative Analysis}
We qualitatively analyze our method on three datasets: Aqua, StrategyQA, and Object tracking. Table \ref{table:qualitative_analysis} exhibits the result of qualitative analysis. These results reveal that the persona is dynamically generated based on the question from the LLM, stating different personas (i.e., Mathematician, Gift Exchange Analyst) for different arithmetic questions. Running the proposed evaluator eventually gives a consistent verdict, mainly deriving the correct output for the given question. This meticulous process aids the framework in mitigating position bias inside the LLM when using it as an evaluator, increasing the performance of the framework.

\section{Conclusion}
In this paper, we propose Jekyll \& Hyde, a novel framework that solves the reasoning problem by ensembling personated and neutral perspectives. 
Evaluations across twelve representative reasoning benchmark datasets present that our framework surpasses both cases when the persona is assigned or not on most datasets.
In addition, our framework's novel method of mitigating position bias has shown better performance when compared with existing methods. 
These results highlight the impact of utilizing LLM with and without persona to improve performance in reasoning tasks. 
Overall, this work sets the initial stage for further investigation in combining solutions from different viewpoints within the LLM community, a promising research direction for improving reasoning abilities.

\newpage
\section*{Limitations}
Although existing methods require additional computation cost and Jekyll \& Hyde is comparable in efficiency to them, our method still requires more computation cost than single perspective LLMs since it runs the LLM at least twice for a single instance. Of course, users can leverage Jekyll \& Hyde practically by setting the maximum attempt of the model into 2, which still outperforms using a single perspective LLM. In addition, the performance of Jekyll \& Hyde is upper-bounded since questions that both neutral and persona LLMs answered incorrectly can not be answered correctly using Jekyll \& Hyde. These aspects of exploration are left to future works.


\section*{Acknowledgements}
This work was supported by Institute of Information \& communications Technology Planning \& Evaluation (IITP) grant funded by the Korea government (MSIT) [No.RS-2022-II220184, Development and Study of AI Technologies to Inexpensively Conform to Evolving Policy on Ethics \& No.RS-2021-II211343, Artificial Intelligence Graduate School Program (Seoul National University) \& No.RS-2021-II212068, Artificial Intelligence Innovation Hub (Artificial Intelligence Institute, Seoul National University)]. K. Jung is with ASRI, Seoul National University, Korea. The Institute of Engineering Research at Seoul National University provided research facilities for this work.

\bibliography{custom}

\begin{thebibliography}{22}
\expandafter\ifx\csname natexlab\endcsname\relax\def\natexlab#1{#1}\fi

\bibitem[{AI@Meta(2024)}]{llama3modelcard}
AI@Meta. 2024.
\newblock \href {https://github.com/meta-llama/llama3/blob/main/MODEL_CARD.md} {Llama 3 model card}.

\bibitem[{Cobbe et~al.(2021)Cobbe, Kosaraju, Bavarian, Chen, Jun, Kaiser, Plappert, Tworek, Hilton, Nakano et~al.}]{gsm8k}
Karl Cobbe, Vineet Kosaraju, Mohammad Bavarian, Mark Chen, Heewoo Jun, Lukasz Kaiser, Matthias Plappert, Jerry Tworek, Jacob Hilton, Reiichiro Nakano, et~al. 2021.
\newblock Training verifiers to solve math word problems.
\newblock \emph{arXiv preprint arXiv:2110.14168}.

\bibitem[{Deshpande et~al.(2023)Deshpande, Murahari, Rajpurohit, Kalyan, and Narasimhan}]{deshpande2023toxicity}
Ameet Deshpande, Vishvak Murahari, Tanmay Rajpurohit, Ashwin Kalyan, and Karthik Narasimhan. 2023.
\newblock Toxicity in chatgpt: Analyzing persona-assigned language models.
\newblock In \emph{Findings of the Association for Computational Linguistics: EMNLP 2023}, pages 1236--1270.

\bibitem[{Geva et~al.(2021)Geva, Khashabi, Segal, Khot, Roth, and Berant}]{strategy}
Mor Geva, Daniel Khashabi, Elad Segal, Tushar Khot, Dan Roth, and Jonathan Berant. 2021.
\newblock \href {https://doi.org/10.1162/tacl_a_00370} {Did aristotle use a laptop? a question answering benchmark with implicit reasoning strategies}.
\newblock \emph{Transactions of the Association for Computational Linguistics}, 9:346--361.

\bibitem[{Gupta et~al.(2023)Gupta, Shrivastava, Deshpande, Kalyan, Clark, Sabharwal, and Khot}]{gupta2023bias}
Shashank Gupta, Vaishnavi Shrivastava, Ameet Deshpande, Ashwin Kalyan, Peter Clark, Ashish Sabharwal, and Tushar Khot. 2023.
\newblock Bias runs deep: Implicit reasoning biases in persona-assigned llms.
\newblock \emph{arXiv preprint arXiv:2311.04892}.

\bibitem[{Hosseini et~al.(2014)Hosseini, Hajishirzi, Etzioni, and Kushman}]{addsub}
Mohammad~Javad Hosseini, Hannaneh Hajishirzi, Oren Etzioni, and Nate Kushman. 2014.
\newblock \href {https://doi.org/10.3115/v1/D14-1058} {Learning to solve arithmetic word problems with verb categorization}.
\newblock In \emph{Proceedings of the 2014 Conference on Empirical Methods in Natural Language Processing ({EMNLP})}, pages 523--533, Doha, Qatar. Association for Computational Linguistics.

\bibitem[{Kojima et~al.(2022)Kojima, Gu, Reid, Matsuo, and Iwasawa}]{kojima2022large}
Takeshi Kojima, Shixiang~Shane Gu, Machel Reid, Yutaka Matsuo, and Yusuke Iwasawa. 2022.
\newblock Large language models are zero-shot reasoners.
\newblock \emph{Advances in neural information processing systems}, 35:22199--22213.

\bibitem[{Koncel-Kedziorski et~al.(2015)Koncel-Kedziorski, Hajishirzi, Sabharwal, Etzioni, and Ang}]{singleeq}
Rik Koncel-Kedziorski, Hannaneh Hajishirzi, Ashish Sabharwal, Oren Etzioni, and Siena~Dumas Ang. 2015.
\newblock \href {https://doi.org/10.1162/tacl_a_00160} {Parsing algebraic word problems into equations}.
\newblock \emph{Transactions of the Association for Computational Linguistics}, 3:585--597.

\bibitem[{Kong et~al.(2024)Kong, Zhao, Chen, Li, Qin, Sun, Zhou, Wang, and Dong}]{kong2024better}
Aobo Kong, Shiwan Zhao, Hao Chen, Qicheng Li, Yong Qin, Ruiqi Sun, Xin Zhou, Enzhi Wang, and Xiaohang Dong. 2024.
\newblock Better zero-shot reasoning with role-play prompting.
\newblock In \emph{Proceedings of the 2024 Conference of the North American Chapter of the Association for Computational Linguistics: Human Language Technologies (Volume 1: Long Papers)}, pages 4099--4113.

\bibitem[{Li et~al.(2023)Li, Wang, Ma, Wu, Wang, Gao, and Liu}]{li2023split}
Zongjie Li, Chaozheng Wang, Pingchuan Ma, Daoyuan Wu, Shuai Wang, Cuiyun Gao, and Yang Liu. 2023.
\newblock Split and merge: Aligning position biases in large language model based evaluators.
\newblock \emph{arXiv preprint arXiv:2310.01432}.

\bibitem[{Ling et~al.(2017)Ling, Yogatama, Dyer, and Blunsom}]{aqua}
Wang Ling, Dani Yogatama, Chris Dyer, and Phil Blunsom. 2017.
\newblock \href {https://doi.org/10.18653/v1/P17-1015} {Program induction by rationale generation: Learning to solve and explain algebraic word problems}.
\newblock In \emph{Proceedings of the 55th Annual Meeting of the Association for Computational Linguistics (Volume 1: Long Papers)}, pages 158--167, Vancouver, Canada. Association for Computational Linguistics.

\bibitem[{OpenAI(2023)}]{openai2023gpt4}
OpenAI. 2023.
\newblock \href {http://arxiv.org/abs/2303.08774} {Gpt-4 technical report}.

\bibitem[{Patel et~al.(2021)Patel, Bhattamishra, and Goyal}]{svamp}
Arkil Patel, Satwik Bhattamishra, and Navin Goyal. 2021.
\newblock \href {https://doi.org/10.18653/v1/2021.naacl-main.168} {Are {NLP} models really able to solve simple math word problems?}
\newblock In \emph{Proceedings of the 2021 Conference of the North American Chapter of the Association for Computational Linguistics: Human Language Technologies}, pages 2080--2094, Online. Association for Computational Linguistics.

\bibitem[{Roy and Roth(2015)}]{multiarith}
Subhro Roy and Dan Roth. 2015.
\newblock \href {https://doi.org/10.18653/v1/D15-1202} {Solving general arithmetic word problems}.
\newblock In \emph{Proceedings of the 2015 Conference on Empirical Methods in Natural Language Processing}, pages 1743--1752, Lisbon, Portugal. Association for Computational Linguistics.

\bibitem[{Shanahan et~al.(2023)Shanahan, McDonell, and Reynolds}]{shanahan2023role}
Murray Shanahan, Kyle McDonell, and Laria Reynolds. 2023.
\newblock Role play with large language models.
\newblock \emph{Nature}, 623(7987):493--498.

\bibitem[{Srivastava et~al.(2022)Srivastava, Rastogi, Rao, Shoeb, Abid, Fisch, Brown, Santoro, Gupta, Garriga-Alonso et~al.}]{srivastava2022beyond}
Aarohi Srivastava, Abhinav Rastogi, Abhishek Rao, Abu Awal~Md Shoeb, Abubakar Abid, Adam Fisch, Adam~R Brown, Adam Santoro, Aditya Gupta, Adri{\`a} Garriga-Alonso, et~al. 2022.
\newblock Beyond the imitation game: Quantifying and extrapolating the capabilities of language models.
\newblock \emph{arXiv preprint arXiv:2206.04615}.

\bibitem[{Talmor et~al.(2019)Talmor, Herzig, Lourie, and Berant}]{csqa}
Alon Talmor, Jonathan Herzig, Nicholas Lourie, and Jonathan Berant. 2019.
\newblock \href {https://doi.org/10.18653/v1/N19-1421} {{C}ommonsense{QA}: A question answering challenge targeting commonsense knowledge}.
\newblock In \emph{Proceedings of the 2019 Conference of the North {A}merican Chapter of the Association for Computational Linguistics: Human Language Technologies, Volume 1 (Long and Short Papers)}, pages 4149--4158, Minneapolis, Minnesota. Association for Computational Linguistics.

\bibitem[{Wang et~al.(2023)Wang, Li, Chen, Cai, Zhu, Lin, Cao, Liu, Liu, and Sui}]{wang2023large}
Peiyi Wang, Lei Li, Liang Chen, Zefan Cai, Dawei Zhu, Binghuai Lin, Yunbo Cao, Qi~Liu, Tianyu Liu, and Zhifang Sui. 2023.
\newblock Large language models are not fair evaluators.
\newblock \emph{arXiv preprint arXiv:2305.17926}.

\bibitem[{Wang et~al.(2022)Wang, Wei, Schuurmans, Le, Chi, Narang, Chowdhery, and Zhou}]{wang2022self}
Xuezhi Wang, Jason Wei, Dale Schuurmans, Quoc Le, Ed~Chi, Sharan Narang, Aakanksha Chowdhery, and Denny Zhou. 2022.
\newblock Self-consistency improves chain of thought reasoning in language models.
\newblock \emph{arXiv preprint arXiv:2203.11171}.

\bibitem[{Wei et~al.(2022)Wei, Wang, Schuurmans, Bosma, ichter, Xia, Chi, Le, and Zhou}]{chain}
Jason Wei, Xuezhi Wang, Dale Schuurmans, Maarten Bosma, brian ichter, Fei Xia, Ed~Chi, Quoc~V Le, and Denny Zhou. 2022.
\newblock \href {https://proceedings.neurips.cc/paper_files/paper/2022/file/9d5609613524ecf4f15af0f7b31abca4-Paper-Conference.pdf} {Chain-of-thought prompting elicits reasoning in large language models}.
\newblock In \emph{Advances in Neural Information Processing Systems}, volume~35, pages 24824--24837. Curran Associates, Inc.

\bibitem[{Zheng et~al.(2024)Zheng, Chiang, Sheng, Zhuang, Wu, Zhuang, Lin, Li, Li, Xing et~al.}]{zheng2024judging}
Lianmin Zheng, Wei-Lin Chiang, Ying Sheng, Siyuan Zhuang, Zhanghao Wu, Yonghao Zhuang, Zi~Lin, Zhuohan Li, Dacheng Li, Eric Xing, et~al. 2024.
\newblock Judging llm-as-a-judge with mt-bench and chatbot arena.
\newblock \emph{Advances in Neural Information Processing Systems}, 36.

\bibitem[{Zheng et~al.(2023)Zheng, Pei, and Jurgens}]{zheng2023helpful}
Mingqian Zheng, Jiaxin Pei, and David Jurgens. 2023.
\newblock Is" a helpful assistant" the best role for large language models? a systematic evaluation of social roles in system prompts.
\newblock \emph{arXiv preprint arXiv:2311.10054}.

\end{thebibliography}
\newpage
\appendix
\section{Prompt Design}
\label{sec:prompt design}
In Jekyll \& Hyde, we leverage three types of LLMs, namely \textbf{Persona Generator}, \textbf{Solver}, and \textbf{Evaluator}. Since each LLM has different roles, they also have different persona designs. Table \ref{tab:persona template}, \ref{tab:eval_template} shows the Persona Generator and Evaluator prompt, respectively.
These prompt designs are followed by \cite{zheng2024judging}, and we manually revise them to give better instructions for all LLM baselines.

\begin{table}[H]
\small
\begin{tcolorbox}

\textcolor[rgb]{0.8,0,0}{SystemMessage:}

You have a special ability in giving job recommendations that could sufficiently solve the given problem. \\

\textcolor[rgb]{0.8,0,0}{HumanMessage:}

This is the user's question: \textcolor[rgb]{0,0,0.9}{\{input\}} \\

According to the question, recommend a job that can sufficiently solve the user's question. Here are some rules you need to follow: \\

1. give a description of the job in JSON format with the following keys:

- job: a specific job name \\

2. Do not give any reasons or preambles about your response \\

Output:












\end{tcolorbox}
\caption{The template for persona generator with one slot \textcolor[rgb]{0,0,0.9}{\{input\}}. Based on the given template, the persona generator yields a unified occupation name (e.g. \textit{Math teacher})}
\vspace{-0.45cm}
\label{tab:persona template}
\end{table}

\begin{table}[t]
\small
\begin{tcolorbox}

Please act as an impartial judge and evaluate the quality of the responses provided by two AI assistants to the user question displayed below.

Your evaluation should ONLY consider correctness. You will be given assistant A’s answer, and assistant B’s answer.

Your job is to evaluate which assistant’s answer is better. You should independently solve the user question step-by-step first

Then compare both assistants’ answers with your answer. Identify and correct any mistakes.

Based on the given two solutions for the following question, you need to choose the best solution based on their explanation and answer

First, solve the problem step by step, and then identify errors and flaws from the given solutions if needed. \\

Please note that:

1. Avoid any position biases and ensure that the order in which the responses were presented does not influence your decision.

2. Do not allow the length of the responses to influence your evaluation.

3. Do not favor certain names of the assistants. Be as objective as possible.

4. Give reason for your choice between two solution.

5. You must output your final verdict by strictly following this format:
"[[A]]" if assistant A is better, and "[[B]]" if assistant B is better \\

This is your user's question: \textcolor[rgb]{0,0,0.9}{\{question\}} \\

assistant A's answer: \textcolor[rgb]{0,0,0.9}{\{assistantA\_answer\}}

assistant A's explanation: \textcolor[rgb]{0,0,0.9}{\{assistantA\_explanation\}} \\

assistant B's answer: \textcolor[rgb]{0,0,0.9}{\{assistantB\_answer\}}

assistant B's explanation: \textcolor[rgb]{0,0,0.9}{\{assistantB\_explanation\}} \\

Now, begin!

Final verdict:













\end{tcolorbox}
\caption{The evaluation template with five slots (\textcolor[rgb]{0,0,0.9}{\{question\}, \{assistantA\_answer\}, \{assistantA\_explanation\}, \{assistantB\_answer\}, and \{assistantB\_explanation\}}). The final verdict output [[A]] or [[B]]}
\label{tab:eval_template}
\end{table}

\section{Solver mechanism}
\label{sec:solver mechanism}
\begin{figure}[t]
\centering
\includegraphics[width=1.0\linewidth]{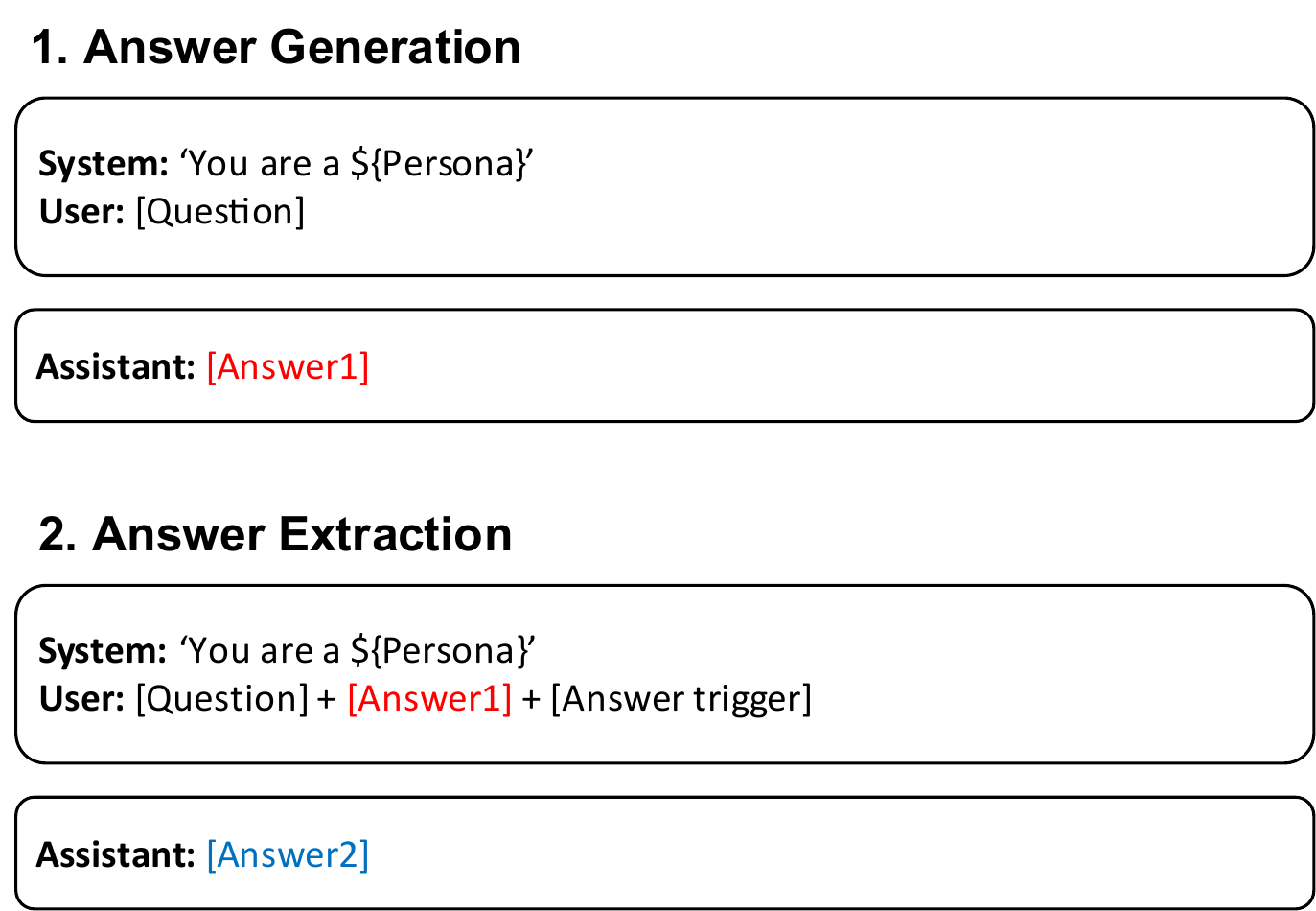}
\caption{ an entire process of how Solver works}
\label{fig:solver_mechanism}
\vspace{-0.4cm}
\end{figure}

When running the LLM under the zero-shot setting, the response is not fixed in a certain format. To extract the answer from the response, we follow the technique of Zero-Shot CoT\cite{kojima2022large}. In detail, the technique consists of two steps, which first generates the response from the LLM based on role-playing prompting and the given question. Then, we concatenate the question, response from the previous step, and an answer trigger together and input them to the LLM, computing the extracting the final answer from the response. The entire progress is shown in figure \ref{fig:solver_mechanism}. The answer trigger sentences for various datasets are depicted in Table \ref{table:trigger}.

\begin{table*}[t]
\centering
\begin{tabular}{ll} 
\toprule
Answer
  Format & Answer
  Trigger                               \\ 
\midrule
arabic number     & Therefore,
  the answer (arabic numerals) is   \\
option (A-E)    & Therefore,
  among A through E, the answer is  \\
option (A-C)    & Therefore,
  among A through C, the answer is  \\
yes or no       & Therefore,
  the answer (Yes or No) is         \\
string          & Therefore,
  the final answer is               \\
\bottomrule
\end{tabular}
\caption{Answer trigger sentences for various answer formats.}
\label{table:trigger}
\end{table*}

\section{Dataset Details}
\label{sec:dataset details}
In this section, we briefly introduce twelve datasets spanning four categories below. Specific details are shown in Table \ref{tab:dataset_description}
\paragraph{Arithmetic.} We leveraged the following six datasets: MultiArith, GSM8K, AddSub, AQUA, SingleEq, and SVAMP. All questions in these datasets include a particular scenario and require reasoning based on mathematical knowledge.
\paragraph{Commonsense Reasoning.} We employ CommonsenseQA and StrategyQA. Both of them require reasoning based on common sense.
\paragraph{Symbolic Reasoning.} we utilize Last letter concatenation and Coin Flip. Last Letter Concatenation demands concatenation of the last letter of the given four words. Coin Flip gives a sequence of operations to flip a coin and asks for the final state of the coin. We utilized these two datasets following the approach of \citet{kojima2022large}.
\paragraph{Other Reasoning Tasks.} We use Date Understanding and Tracking Shuffled Objects from Bigbench\cite{srivastava2022beyond}. Date Understanding requires date calculations. Tracking Shuffled Objects gives a sequence of object substitution operations and then asking the final location of a certain object.
\begin{table*}[t]
\centering
\scalebox{0.82}{
\setlength{\tabcolsep}{20pt}
\begin{tabular}{lcccc}
\toprule
Dataset &Answer Format & $N_{q}$ &$L_{q}$ &License \\\midrule
SingleEq &arabic number &508 &27.4 &No License \\
AddSub &arabic number &395 &31.5  &Unspecified \\
MultiArith &arabic number &600 &31.8  &Unspecified \\
GSM8K &arabic number &1319 &46.9  &MIT License \\
AQUA &option (A-E) &254 &51.9  &Apache-2.0 \\
SVAMP &arabic number &1000 &31.8  &MIT License \\
CommonsenseQA &option (A-E) &1221 &27.8  &Unspecified \\
StrategyQA &yes or no &2290 &9.6  &Apache-2.0 \\
Date Understanding &option (A-F) &369 &35.0  &Apache-2.0 \\
Object Tracking &option (A-C) &750 &91.1 &Apache-2.0 \\
Last Letters &string &500 &15.0  &- \\
Coin Flip &yes or no &500 &37.0  &- \\
\bottomrule
\end{tabular}}
\caption{Relevant information of 12 datasets. $N_{q}$ denotes the number of questions in each dataset. $L_{q}$ denotes the average words of questions in each dataset.}
\label{tab:dataset_description}
\end{table*}

\section{Confusion matrix for other datasets}
\begin{table*}[]
\centering
\small
\begin{tabular}{cc|c|c||c|c|c||c|c|c}

\toprule
\multirow{2}{*}{\makecell{Method}}&\multicolumn{9}{c}{\makecell{Persona Solver \\ (w/ Persona)}} \\
\cmidrule{2-10}
& \multicolumn{3}{c}{\makecell{StrategyQA}} & \multicolumn{3}{c}{\makecell{Coin Flip}} & \multicolumn{3}{c}{\makecell{Object Tracking}} \\
\midrule
\multirow{4}{*}{\makecell{Neutral Solver \\ (w/o Persona)}} &  & Wrong & Right &  & Wrong & Right &  & Wrong & Right \\
\cmidrule{2-10}
& Wrong & 19.39\% & 12.31\% & Wrong & 4.60\% & 4.00\% & Wrong & 46.67\% & 18.13\% \\
\cmidrule{2-10}
&Right& \textcolor{red}{\textbf{10.31\%}} & 57.99\% & Right & \textcolor{red}{\textbf{18.00\%}} & 73.40\% & Right & \textcolor{red}{\textbf{12.93\%}} & 22.27\% \\
\bottomrule
\end{tabular}
\caption{Confusion matrix between Neutral Solver (w/o Persona) and its Persona Solver (w/ Persona) on StrategyQA dataset.}
\label{table: confusion matrix other datasets}
\end{table*}
\label{sec:all_confusion_matrix}
As shown in Table \ref{table: confusion matrix aqua}, we reveal that some of the questions are correctly answered with LLMs without role-playing prompting, while getting wrong when using LLM with role-playing prompting. Here, we provide the result of a confusion matrix for other datasets, namely the StrategyQA, Coin Flip, and Object Tracking datasets. Table \ref{table: confusion matrix other datasets} exhibits the confusion matrix for each dataset respectively.

\section{Impact of prompt design}
\label{sec:impact_of_prompt_design}
This section introduces the default prompt design for persona LLM. While there are a lot of variations in prompts, we are the first to compare the impact of prompt designs for LLM-generated role-playing prompts according to the best of our knowledge. Hence, we conducted three different prompts and computed the performance of each prompt with GPT-3.5-turbo using the Aqua dataset. Table \ref{table:prompt_design} shows different forms of prompts and their performance. The result reveals that using a single persona acquires the optimal performance in persona LLM, thereby outperforming other settings in Jekyll \& Hyde.

\begin{table*}[]
\resizebox{1.0\textwidth}{!}
{
\centering
\begin{tabular}{@{}c|p{10cm}|c|c@{}}
\toprule
form & prompt & AQUA & Accuracy ($\uparrow$) \\
\midrule
persona & You are a [$persona$] & Persona & \textbf{65.75}\\
& & Jekyll $\&$ Hyde & \textbf{69.68} \\
\midrule
persona + task description & You are a [$persona$]. Your task is to solve the given math question and come up with a correct answer. & Persona & 62.99 \\
& & Jekyll $\&$ Hyde & 68.11 \\
\midrule
task description & Your task is to solve the given math question and come up with a correct answer. & Persona & 65.35 \\
& & Jekyll $\&$ Hyde & 68.90 \\
\bottomrule
\end{tabular}
}
\caption{
\textbf{Performance of different prompt designs} Among different types of prompt design, using only persona for the prompt obtains the highest performance in Persona LLM and Jekyll \& Hyde, thereby setting it as a default prompt for our Persona LLM.
}
\label{table:prompt_design}
\end{table*}

\section{Implementation details for Portia and MEC+BPC}
\label{sec:Portia and MEC+BPC implementation}
In section \ref{section:position_bias_mitigation}, we conduct an experiment to compare the performance of mitigating position bias. Here, we employed two existing methods, specifically Portia and MEC+BPC. 
\paragraph{Portia} is introduced by \citet{li2023split}, which mitigates position biases by slicing each given response into chunks and putting them alternately into the prompt, mitigating the information of the order between the given responses. We implemented this method by slicing the given response into chunks with fixed lengths and then inserting them alternately into the evaluation prompt.
\paragraph{MEC+BPC} is introduced by \citet{wang2023large} to mitigate position bias in the LLM Evaluator. It utilizes two evaluation prompts with differently ordered sequences (in forward and reverse orders) of the response. This method executes each evaluation prompt to estimate the scores of two responses, respectively.
After deriving scores for each response, the final scores of each response are aggregated and computed by averaging scores for the two sequences of solutions, respectively. We implemented MEC+BPC by preparing two evaluation prompts for the two sequences. Then, we ran the model and computed the score for each response. The model is run three times for robust answer generation, and the average of the scores is computed.

\section{Settings for the number of self-consistency of the base, persona LLMs}
\label{sec:settings for sc}
In table \ref{table:compare_sc}, we executed Jekyll \& Hyde for each model and calculated the average number of LLM executions per instance.
In order to compare the performance of the Base and Persona LLMs under the condition of executing the LLM the same number as Jekyll \& Hyde, we ensured self-consistency for both LLMs.
Specifically, we executed Jekyll \& Hyde for both datasets and computed the average number of LLM executions for a single instance of each dataset.
Afterward, the number of self-consistency for Base + voting is determined as the ceiling of the average executions for Jekyll \& Hyde. In the case of Persona + voting, given an average number of LLM executions $n$, we determined the number of self-consistency $k$ following the formula below:
\begin{equation}
\small
    k = \begin{cases}
        2 \cdot \lfloor \frac{n}{2} \rfloor & \text{if $\lceil n \rceil \div 2 = 0$ } \\
        2 \cdot (\lfloor \frac{n}{2} \rfloor + 1) & \text{if $\lceil n \rceil \div 2 = 1$ or $n = 4$}
    \end{cases}
\normalsize
\end{equation}
since Persona requires two times inference (Persona Generator + Persona Solver), we incremented the self-consistency iterations if the number is odd. When $n$ is 4, it means that Persona yields two outputs, leading it impossible to find the most frequent answer if two outputs are different.

\section{Full comparison between self-consistency and Jekyll \& Hyde}
\label{section:full experiment of self-consistency}
Table \ref{table:compare_sc_org} reveals the performance of three different models utilizing Jekyll \& Hyde and self-consistency using Base LLM and Persona LLM.

\begin{table}[h]
\centering
\small
\resizebox{1.0\linewidth}{!}{
\begin{tabular}{ccccc}
\toprule
 \multirow{2}{*}{Model} & \multirow{2}{*}{Datasets} & \multirow{2}{*}{Methods} & \multirow{2}{*}{Accuracy ($\uparrow$)} & \multirow{2}{*}{\makecell{Average \\ LLM runs ($\downarrow$)}} \\
 & & & & \\
\midrule
\multirow{6}{*}{GPT-4} & \multirow{3}{*}{AQuA} & Base + voting & 70.87 & 4 \\
& & Persona + voting & 73.23 & 6 \\
 \cmidrule{3-5}
 & & Jekyll \& Hyde & \textbf{76.90} & 3.81 \\
\cmidrule{2-5}
& \multirow{3}{*}{Object} & Base + voting & 46.00 & 5 \\
& & Persona + voting & 59.20 & 6 \\
\cmidrule{3-5}
& & Jekyll \& Hyde & \textbf{61.69} & 4.14 \\
\midrule
\multirow{6}{*}{GPT-3.5-turbo} & \multirow{3}{*}{AQuA} & Base + voting & 66.14 & 5 \\
& & Persona + voting & 66.53 & 6 \\
 \cmidrule{3-5}
 & & Jekyll \& Hyde & \textbf{67.98} & \textbf{4.35} \\
\cmidrule{2-5}
& \multirow{3}{*}{Object} & Base + voting & 34 & 5 \\
& & Persona + voting & 33.73 & 6 \\
\cmidrule{3-5}
& & Jekyll \& Hyde & \textbf{42.22} & \textbf{4.30} \\
\midrule
\multirow{6}{*}{\makecell{llama3 \\ (8b)}} & \multirow{3}{*}{AQuA} & Base + voting & \textbf{60.63} & 5 \\
& & Persona + voting & 59.06 & 6 \\
 \cmidrule{3-5}
 & & Jekyll \& Hyde & 54.07 & \textbf{4.96} \\
\cmidrule{2-5}
& \multirow{3}{*}{Object} & Base + voting & \textbf{49.87} & 5 \\
& & Persona + voting & 47.07 & 6 \\
\cmidrule{3-5}
& & Jekyll \& Hyde & 48.58 & \textbf{4.59} \\
\bottomrule
\end{tabular}}
\caption{\textbf{Comparison of performance between Jekyll \& Hyde, Base with self-consistency, and Persona with self-consistency} Base + sc indicates using base llm with self-consistency, and Persona + sc indicates using persona LLM with self-consistency.}
\vspace{-15pt}
\label{table:compare_sc_org}
\end{table}

\section{Full experiment for comparing methods of positional bias mitigation}
\label{section:full experiment of position bias}
Table \ref{table:pos_bias_org} shows the performance of three different position bias mitigation methods for 12 datasets.
Portia is implemented by dividing the given solution into three chunks, each with the same number of tokens. MEC+BPC is implemented by generating scores ranging from 1 to 10 three times for each solution. The final solution is determined by comparing the average score of each solution. The result exhibits that utilizing the Jekyll \& Hyde evaluator achieves optimal performance across most datasets.

\begin{table*}[h]
\centering
\small
\resizebox{1.0\linewidth}{!}{
\begin{tabular}{cccccccc}
    \toprule
    \hspace{0.5em}Models & Method & AddSub & AQuA & SingleEq & SVAMP & Coin & Date \\ 
    \midrule
    \multirow{5}{*}{GPT-4} & Oracle Evaluator & 97.72 & 81.10 & 99.41 & 95.20 & 88.80 & 82.66 \\
    \cmidrule{2-8}
    & Portia & 97.47 & 74.41 & 98.82 & 91.80 & 74.40 & 80.76 \\
    & MEC+BPC & 97.22 & 74.41 & 98.43 & 91.20 & 74.00 & 79.95 \\
    & \makecell{Jekyll \& Hyde$^{\dag}$} & 97.72 & \textbf{78.35} & 98.43 & 92.20 & 78.20 & 80.22 \\
    \cmidrule{2-8}
    & Jekyll \& Hyde & \textbf{97.72} & 77.56 & \textbf{99.02} & \textbf{92.60} & \textbf{79.80} & \textbf{81.57} \\
    \midrule
    \multirow{5}{*}{GPT-3.5-turbo} & Oracle Evaluator & 95.19 & 74.41 & 99.21 & 87.10 & 60.80 & 80.22 \\
    \cmidrule{2-8}
    & Portia & 91.14  & 62.60 & 98.23 & 81.80 & 57.80 & 72.63  \\
    & MEC+BPC & 89.37 & 62.60 & 97.64 & 80.20  & 57.60 & \textbf{75.61} \\
    & \makecell{Jekyll \& Hyde$^{\dag}$} & 92.15 & 62.60 & 97.83 & 82.50 & 56.60 & 72.63 \\
    \cmidrule{2-8}
    & Jekyll \& Hyde & \textbf{93.16} & \textbf{64.17} & \textbf{98.23} & \textbf{83.00} & \textbf{59.60} & 72.63 \\
    \midrule
    \multirow{5}{*}{\makecell{llama3 \\ (8B)}} & Oracle Evaluator & 92.41 & 63.39 & 96.06 & 90.20 & 99.00 & 84.55 \\
    \cmidrule{2-8}
    & Portia & 88.35 & 51.97 & 93.31 & 86.10 & 96.40 & 78.86 \\
    & MEC+BPC & 88.10 & \textbf{55.91} & 91.73 & 84.50 & 95.40 & \textbf{81.03} \\
    & \makecell{Jekyll \& Hyde$^{\dag}$} & 90.38 & 51.18 & 94.29 & 86.10 & 97.00 & 79.95 \\
    \cmidrule{2-8}
    & Jekyll \& Hyde & \textbf{91.14} & 53.54 & \textbf{95.67} & \textbf{86.80} & \textbf{98.40} & 79.95 \\
    \bottomrule
\end{tabular}}
\caption{
\textbf{Mitigating positional bias.} We report that the LLM Evaluator used for Jekyll \& Hyde outperforms other existing methods in most datasets. Despite the marginal increase when using the LLM Evaluator from Jekyll \& Hyde, the Evaluator aids the LLM to nearly approach the performance of an Oracle Evaluator, which is the optimal performance for the given datasets.}
\label{table:pos_bias_org}
\end{table*}

\section{Hyperparameter settings for GPT-4 and GPT-3.5-turbo}
\label{section:full hyperparameter setting}
\begin{figure}[t]
\centering
\includegraphics[width=1.0\linewidth]{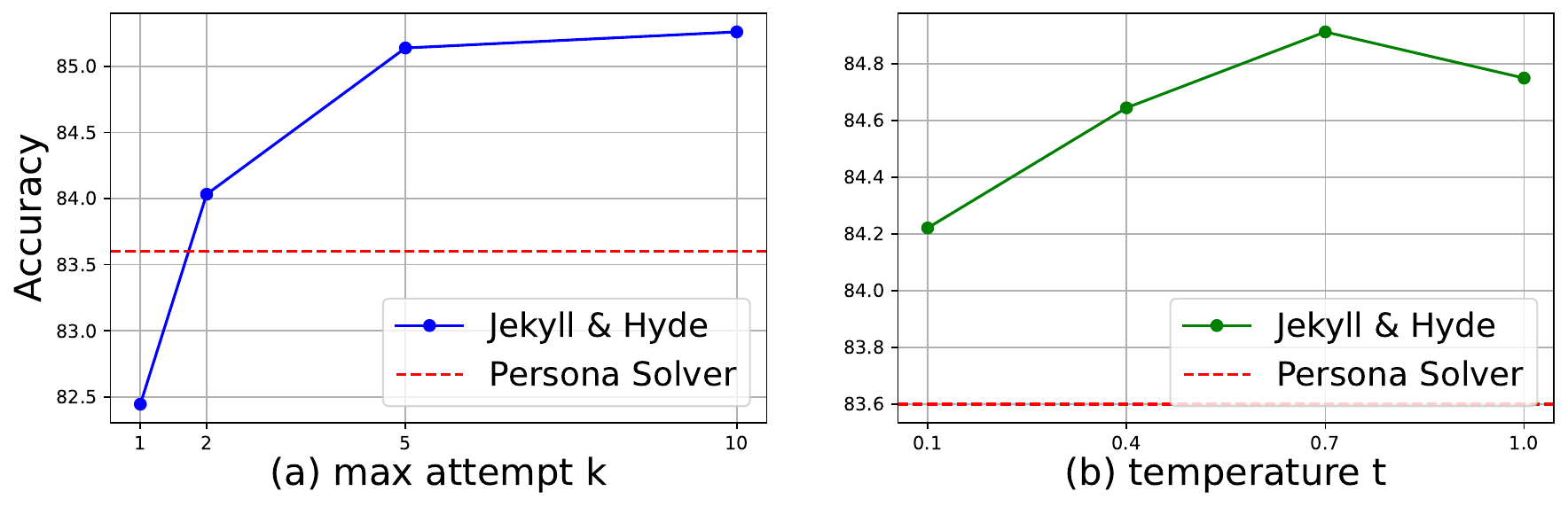}
\caption{\textbf{Hyper-parameters Experiments for gpt-3.5-turbo} 
Variation of averaged accuracy with a (a) various number of max attempt $k$ and (b) temperature of the LLM $\tau$ used in LLM evaluator. X and Y axes correspond to each hyper-parameter setting and accuracy, respectively.
}
\vspace{-0.4cm}
\label{fig:hparams_setting_turbo}
\end{figure}

\begin{figure}[h]
\centering
\includegraphics[width=1.0\linewidth]{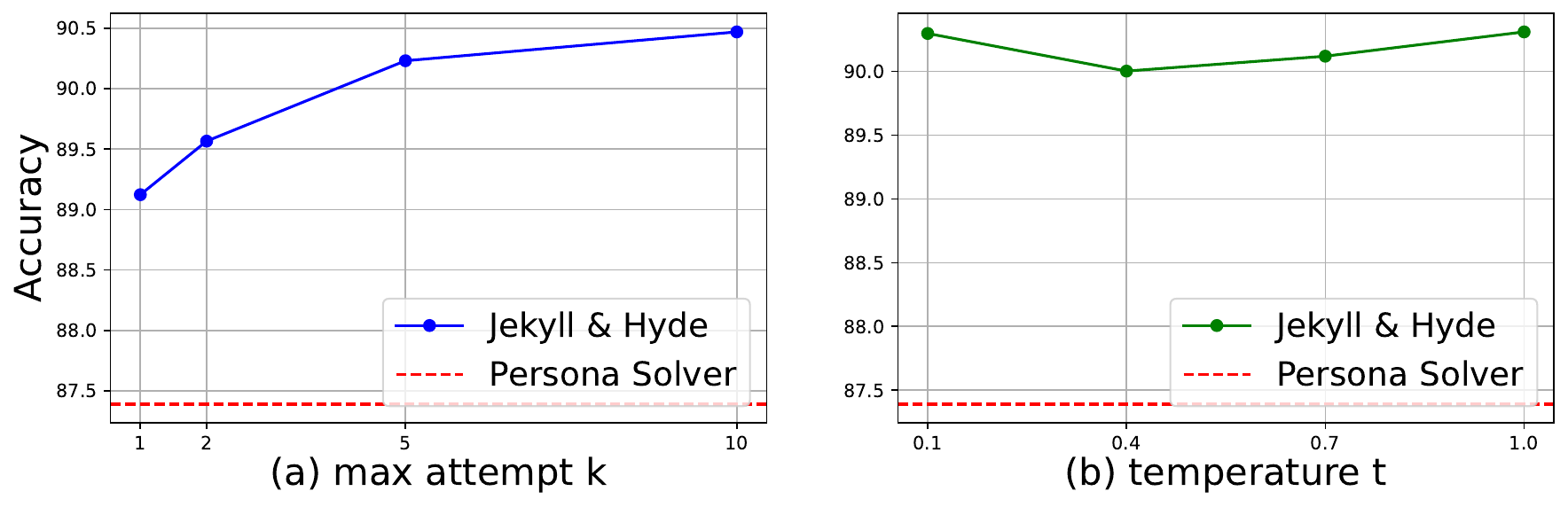}
\caption{\textbf{Hyper-parameters Experiments for GPT-4} 
Variation of averaged accuracy with a (a) various number of max attempt $k$ and (b) temperature of the LLM $\tau$ used in LLM evaluator. X and Y axes correspond to each hyper-parameter setting and accuracy, respectively.
}
\vspace{-0.4cm}
\label{fig:hparams_setting_gpt-4}
\end{figure}

Figures \ref{fig:hparams_setting_turbo}, and \ref{fig:hparams_setting_gpt-4} show the experimental result for the hyperparameter setting. As it shows, GPT-3.5-turbo shows that obtaining 0.7 as a temperature achieves the highest performance among other settings, and GPT-4 reveals that using 0.1 or 1.0 as a temperature yields the highest performance. 
Since using 0.7 as a temperature does not lead the model to a significant performance decline, we determined 0.7 as our default temperature.
Meanwhile, both GPT-3.5-turbo and GPT-4 present that the slope of the graph gradually flattens as the maximum number of attempts increases, leading to performance saturation at a certain performance. Hence we concluded to use 5 as our default max attempt setting considering the trade-off between the performance and the computational cost.


\end{document}